\documentclass[10pt,twocolumn,letterpaper]{article}

\usepackage{times}
\usepackage{epsfig}
\usepackage{graphicx}
\usepackage{amsmath}
\usepackage{amssymb}
\usepackage{booktabs}
\usepackage[dvipsnames]{xcolor}
\usepackage{authblk}
\usepackage[margin=0.6in]{geometry}
\usepackage[square,sort,comma,numbers]{natbib}
\definecolor{mygreen}{rgb}{0,0.7,0}
\usepackage{color, soul}

\usepackage{xcolor}
\usepackage{xspace}
  \usepackage{colortbl}
  \usepackage{arydshln,graphicx,xcolor,array}

\usepackage{algpseudocode}
\usepackage[]{algorithm}
\usepackage[acronym]{glossaries}
\usepackage{array,multirow,multicol}
\usepackage{gensymb}

\usepackage[pagebackref=true,breaklinks=true,letterpaper=true,colorlinks,bookmarks=false]{hyperref}

\makeglossaries{}
\newacronym{wrt}{\textit{w.r.t}}{with respect to}
\newacronym{mAP}{\textit{mAP}}{mean Average Precision}
\newacronym{sota}{SoA}{State-of-the-Art}
\newacronym{fig}{Fig.}{Figure}
\newacronym{tab}{Tab.}{Table}
\newacronym{alg}{Alg.}{Algorithm}
\newacronym{bow}{BoW}{Bag of Words}
\newacronym{cnn}{CNN}{Convolutional Neural Network}
\newacronym{hog}{HoG}{Histogram of Oriented Gradients}
\newacronym{oid}{OID}{Open Images Dataset V4}
\newacronym{sfm}{SfM}{Structure-from-Motion}
\newacronym{slam}{SLAM}{Simultaneous Localization And Mapping}
\newacronym{pnp}{PnP}{Perspective-n-Point}
\newacronym{nms}{NMS}{Non Maximum Suppression}
\newacronym{iou}{IoU}{Intersection-over-Union}
\newacronym{nn}{NN}{Nearest-Neighbor}
\newacronym{sgd}{SGD}{Stochastic Gradient Descent}

\newcommand{\Fig}{Fig.\xspace}

\newcommand{\Sec}{Sec.\xspace}
\newcommand{\Tab}{Tab.\xspace}

\newcommand{\PAR}[1]{\vskip4pt \noindent{\bf #1~}}
\newcommand{\eg}{e.g.\@\xspace}

\newcommand{\example}{\textit{e.g.}\xspace}

\newcommand{\horus}{OGuL\xspace} %

\definecolor{cgreen}{RGB}{26, 110, 53}
\definecolor{cgrass}{RGB}{123, 252, 3}
\definecolor{cbrown}{RGB}{161, 100, 56}
\definecolor{cyellow}{RGB}{237, 187, 36}
\definecolor{cpurple}{RGB}{177, 87, 250}
\definecolor{cpurblue}{RGB}{194, 207, 242}
\definecolor{cgrey}{RGB}{157, 163, 163}
\definecolor{corange}{RGB}{245, 130, 69}
\definecolor{cblue}{RGB}{66, 120, 245}
\definecolor{csky}{RGB}{148, 250, 255}
\definecolor{ccyan}{RGB}{8, 189, 171}
\definecolor{crose}{RGB}{235, 101, 157}
\definecolor{cpink}{RGB}{255, 212, 212}
\definecolor{cred}{RGB}{219, 15, 15}
\definecolor{cdark}{RGB}{0, 0, 0}

\newcommand{\hgreen}[1]{\textcolor{cgreen}{#1}\xspace}

\newcommand{\hblue}[1]{\textcolor{cblue}{#1}\xspace}

\begin{document}

\date{\vspace{-5ex}}
\title{Object-Guided Day-Night Visual Localization in Urban Scenes}

\author[1]{Assia Benbihi}
\author[2]{C{\'e}dric Pradalier}
\author[1]{Ond\v{r}ej Chum}
\affil[1]{VRG, Faculty of Electrical Engineering\\Czech Technical University in Prague}
\affil[2]{IRL 2958 GT-CNRS, Metz, France}

\maketitle

\begin{abstract}
  We introduce Object-Guided Localization (\horus) based on a novel method of
  local-feature matching. Direct matching of local features is sensitive to
  significant changes in illumination. In contrast, object detection often
  survives severe changes in lighting conditions. The proposed method first
  detects semantic objects and establishes correspondences of those objects
  between images. Object correspondences provide local coarse alignment of the
  images in the form of a planar homography. These homographies are
  consequently used to guide the matching of local features. Experiments on
  standard urban localization datasets (Aachen, Extended-CMU-Season,
  RobotCar-Season) show that \horus significantly improves localization results
  with as simple local features as SIFT, and its performance competes with the
  state-of-the-art CNN-based methods trained for day-to-night localization.
\end{abstract}

\section{Introduction}

Visual localization is a key step enabling modern technologies, such as autonomous driving or %
augmented reality. %
Classical approaches to visual localization exploit the Structure-from-Motion (SfM) pipeline. In this setup, local features are detected and described
by a high-dimensional appearance descriptor in each image independently. Tentative correspondences of the local features are established based on the similarity of their descriptors. In the final step of the SfM pipeline, scene geometry and camera poses are robustly estimated from the tentative matches.

\begin{figure}[thb]
 \centering
 \includegraphics[width=\linewidth]{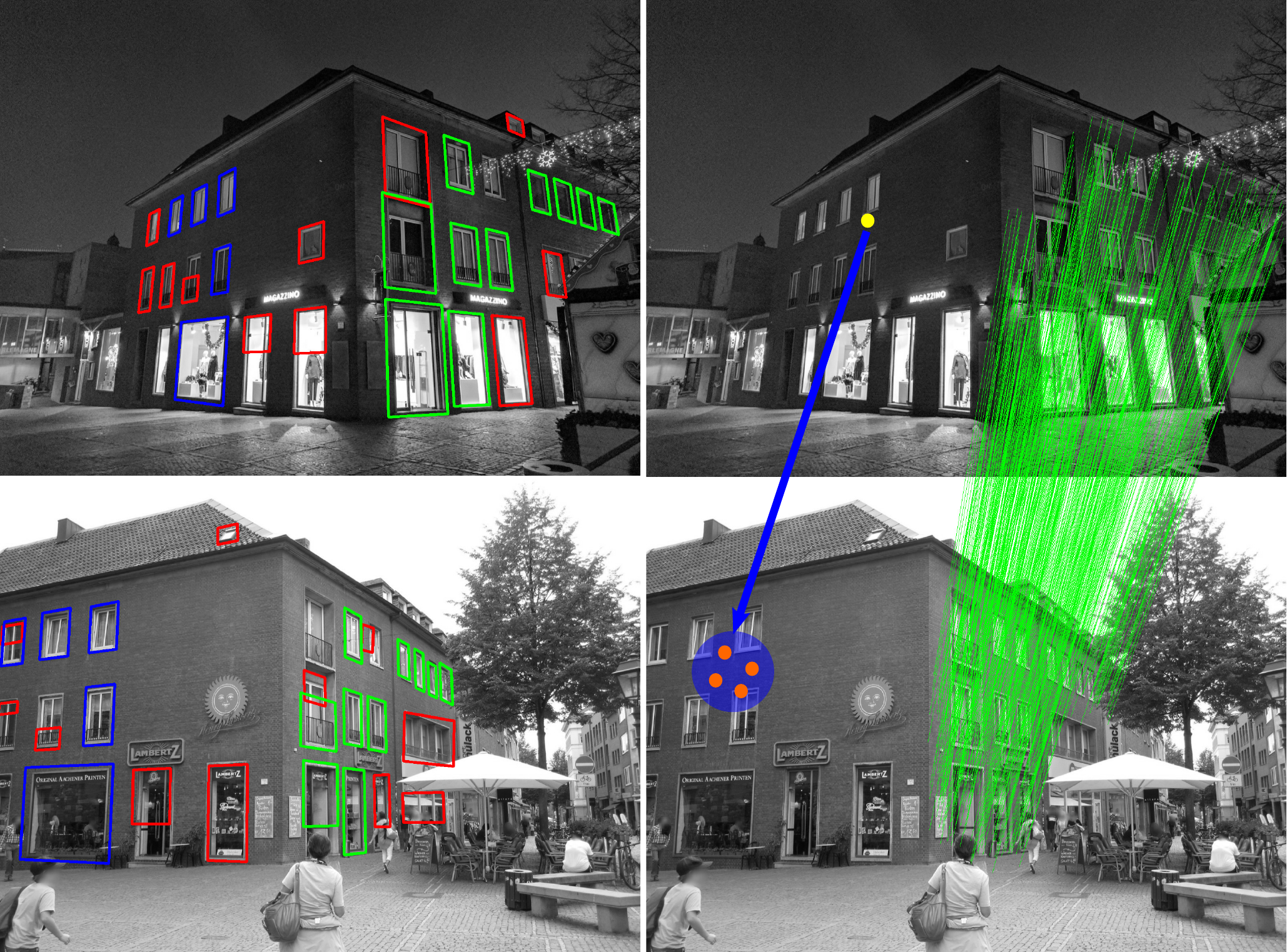}
  \caption{Matching of local features is guided by the coarse geometry of matching
  objects. Left: Bounding boxes of semantic objects detected in the images. Two sets of corresponding objects (related by a planar homography) were detected, shown in blue and green. For the objects in red bounding boxes, no corresponding objects were found in the other image.  Right: Each homography guides the
  matching of local features from the first image (yellow dot) by constraining the search space of possible matches in the second image (orange dots in the blue-circle) - demonstrated on homography relating objects in blue; and a final set of matches demonstrated on homography relating objects in green. }
  \label{fig:fig1}
\end{figure}

This paper targets the task of day-and-night visual localization where severe changes of illumination break the SfM pipeline in the very beginning - at the local features detection and matching stages. 
The negative impact of illumination changes on local feature matching is twofold. First, local feature detectors are unlikely to fire at identical locations. Second, the visual appearance may vary significantly, so that even for features covering the same surface, the extracted descriptors are dissimilar. 
There are recent research directions to alleviate local feature failure by
training detectors and descriptors that are invariant or less sensitive to
severe changes in
illumination~\cite{detone18superpoint,dusmanu2019d2,mishchuk2017working,tian2019sosnet,r2d2}.
However, the detectors and descriptors are not powerful enough to solve the
day-and-night localization problem alone. Therefore we focus on the matching
part of the problem in this paper. 

Urban scenes are a distinct type of man-made environment. On one hand, it introduces challenges that make the matching task harder, such as extreme illumination variations.
The appearance changes are not only caused by different directions, colors, and/or intensities of the light; there are parts of the scene that change their appearance completely. This includes elements that become light sources themselves, for example, street lamps, neon signs, or see-through windows.

On the other hand, the structure of urban environments 
exhibits properties that allow certain geometric reasoning %
based on mild assumptions. %
These assumptions include piece-vice planar structures of the environment, where the majority of the planes are vertical facades. 
Many of the man-made structures on the facade have semantic meaning and it is relatively easy to recognize them on the category level, which is independent of the view-point or illumination conditions (\eg windows). 
Another assumption is the large number of facade structures that tend to be aligned with horizontal and vertical directions. 
This allows for reliable estimation of horizontal and vertical vanishing points, which are useful in geometric reasoning.

In this work, we propose the Object-Guided Localization (\horus) pipeline, that exploits the urban structures that survive severe changes in illumination.
In particular, semantic objects and edges are used. The correspondences between semantic objects generate hypotheses of coarse scene alignment, which are used to guide the matching  of local features, see~\Fig\ref{fig:fig1}. %
Edges are used to estimate vanishing points %
and to improve the detection of semantic objects.
For semantic objects, we restrict to the most commonly appearing facade element -- windows. However, other types of planar objects can be used in a straightforwardly.

\begin{figure*}[thb]
 \centering
  \includegraphics[width=0.95\linewidth]{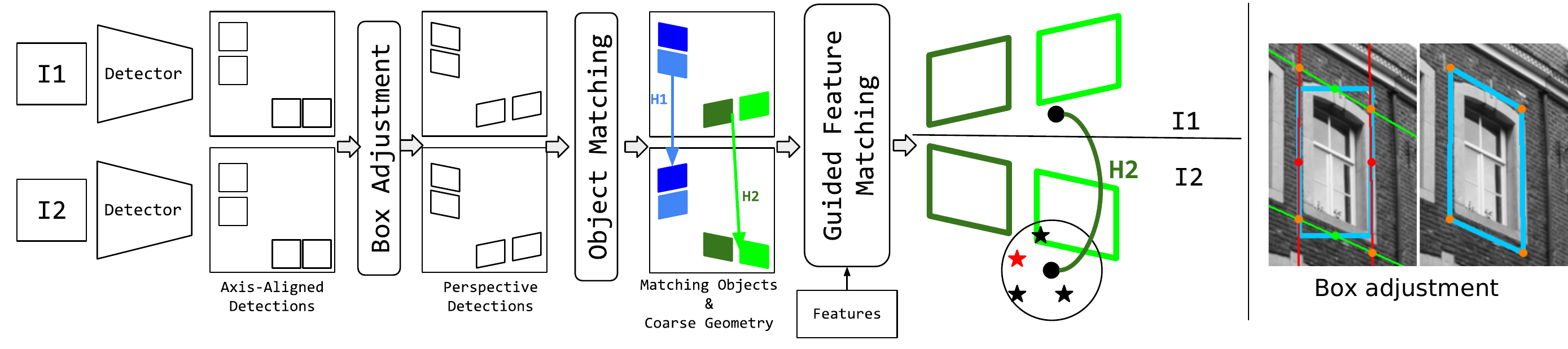}
  \caption{Left: Illustration of the method.
  Semantic objects are detected and their bounding boxes are projectively adjusted using the vanishing points.
  Homography hypotheses registering the largest number of object bounding boxes are estimated.
  The homographies are used to guide the matching of local features. %
  Right: Illustration of the box adjustement. An axis-aligned box is detected in the original image (left) and adjusted
  to be consistent with the vanishing points (right).
  The green / red points are the midpoints of the horizontal / vertical box edges.
  The green / red lines join the midpoints of the box edges with the appropriate vanishing point.
  Intersections of the red and green lines (orange dots) define the  corners of the projective bounding box.
  }
 \label{fig:method_summary}
\end{figure*}

\horus is evaluated on the task of visual localization on three benchmark
datasets, on which it systematically improves day-night localization and maintains
the performance on the day. 
Our contributions are the following: (i) We propose a novel feature matching
guidance strategy based on semantic object pairs that are robust to appearance
variations. (ii) We augment box detection to allow for perspective box
detection without additional training, using geometric properties of the image.
(iii) Our complete pipeline improves day-night localization and manages to
match pairs that are not handled by \gls{sota} methods.

\section{Related Work}
This section provides an overview of local feature matching.

\PAR{Object-Guided Visual Localization.}
The most related work to ours is an object-of-interest-guided visual localization
approach for \emph{indoor} environments~\cite{weinzaepfel2019visual}.
A detector of a fixed set of specific and highly discriminative planar objects of interest is trained. 
For a detected object in a query image, a dense set of point correspondences to
its reference image in the database is established. The final localization is estimated by 2D-to-3D, exploiting the
known 3D position of the object of interest in the environment.
Our method differs in that it detects generic objects and exploits their coarse correspondences to guide the matching of the surrounding pixels. %

\PAR{Coarse-to-Fine Feature Matching} first aligns the images or their parts approximately, then
the matching is refined to achieve higher accuracy. Commonly, such a procedure starts at a lower image resolution
and proceeds to a finer resolution~\cite{liu2008sift,liu2010sift,brox2004high,bruhn2005lucas}.
Modern local features based on CNNs naturally operate at the coarse level as the spatial resolution of the
CNN's output is lower than the resolution of the input image. Various methods to obtain finer matches
were proposed. 

One of such methods is learning-free interpolation, introduced in~\cite{detone18superpoint}. Matches obtained at higher levels of the CNN (lower resolution) are refined at lower levels of the CNN in~\cite{widya2018structure,taira2018inloc}.
In~\cite{darmon2020learning}, receptive fields of corresponding deep features are used to guide the matching of standard local features.
Recently, Patch2Pix~\cite{zhou2021patch2pix} learns to regress a single pixel correspondence, while Loftr~\cite{sun2021loftr} learns a dense
matching of the pixels between corresponding receptive fields.
The coarse matching in these approaches originates from matching at a lower resolution. In our case, the coarse matching is a consequence of the imprecise localization of the detected objects.

\PAR{Guided Matching.} There are two approaches to expanding a restricted set of 
reliable correspondences with either global or local geometric guidance. In the case of global geometry,
the reliable correspondences are used to robustly estimate global geometry and then restrict the search for further
matches to be consistent with it~\cite{Hartley2004}. Various geometric models can be used, \eg epipolar geometry~\cite{shah2015geometry}, or local optical
flow~\cite{maier2016guided,geiger2011stereoscan}. A recent example of such an approach is Ransac-Flow~\cite{shen2020ransac}, where a homography is estimated from the deep features, used
 to warp one image to another and coarsely align them before learning a dense matching.
Another approach starts from a local geometry of the features and gradually extends the neighbourhood ~\cite{ferrari2004simultaneous}.

All the guided matching methods rely on the existence of a sufficient number of correct matches of local features to instantiate the initial geometry. In contrast, the proposed approach
instantiates the initial geometry based on semantic objects, which are more reliably detected under severe photometric transformations.

\PAR{Matching by Global Optimization.}
Another type of guidance relies on a voting scheme that draws matches that
agree the most with each other. This requires the definition of similarity
between two matches that usually embeds their geometric agreement in the form
of translation~\cite{leordeanu2005spectral},
similarity~\cite{puerto2012hierarchical,torresani2008feature}, affine
transform~\cite{chen2013robust,chen2015co}, or optical
flow~\cite{ni2009groupsac}. Matches are then drawn from clustering or graph
optimization algorithms. This is the case for the \gls{sota}
SuperGlue~\cite{sarlin2020superglue} where the similarity is learned using an
attention mechanism.

\PAR{Match Filtering}
is a post-processing step that attempts to filter out the outliers from a given
set of matches. Deep learning methods attempt to classify matches as inliers
or outliers~\cite{yi2018learning,zhang2019learning,brachmann2019neural,sun2020acne}
while hand-crafted methods rely on match similarity~\cite{cho2009feature},
smoothness constraints~\cite{lin2014bilateral,bian2017gms}, or local affine
transform~\cite{cavalli2020handcrafted}. Any of these methods can be applied as a postprocessing step to the proposed method.

\PAR{Semantic Object Alignment.} This line of
work~\cite{rocco2017convolutional,rocco2018end,kim2017fcss} focuses on matching
different object instances of the same category, usually only one object per
image, rather than the same instance under various conditions as in our case.
The goal is to establish dense pixel correspondences between the objects
for applications such as style transfer rather than visual localization.
Most related to our work is the use of object region
proposals to guide dense feature
matching~\cite{ham2016proposal,ham2017proposal,han2017scnet,yang2017object}.
However, these proposals represent only a fragment of the single object to match and
are associated through visual similarity and simple spatial consistency. Instead, \horus detects multiple whole
objects per image and relies on geometry to match them. This
geometric guidance, in the form of a homography, is more informative than the
usual translation and scale parameters used in these works.

\section{Object-Guided Feature Matching}

We propose to generate coarse geometric guidance of local features based on the correspondences of
semantic objects~(\Fig~\ref{fig:method_summary}). First, objects are detected and represented by a perspective
bounding box~(\Sec~\ref{method:object_detection}). Object correspondences are established based on their geometric layout in the images~(\Sec~\ref{method:object_matching}).
A pair of objects defines a local geometry
in the form of a homography derived from the corresponding box corners. This geometric model
is used to guide the neighboring local feature matching
(\Sec~\ref{method:guided_feature_matching}). 
The main advantage of the proposed method is the ability to match local
features even in situations where establishing the feature correspondences
based on appearance only is not possible. %

\subsection{Perspective Object Detection}
\label{method:object_detection}

The first step of the proposed pipeline is the detection of semantic objects. Object detectors commonly output an axis-aligned bounding box, ignoring the perspective distortion.
Since the detected objects are used to estimate a homography transformation, more precise localization is needed.
Two complementary methods exploiting estimated vanishing points are proposed. In the first
one~(\Sec~\ref{method:vp_alignment}), boxes are detected %
and their edges are adjusted so
that they coincide with the vanishing directions of the plane the object lies
on (\Fig~\ref{fig:method_summary}-right). 
The second approach
(\Sec~\ref{method:image_rectification}) rectifies the image before the boxes are
detected so that the planar objects are orthogonal in the rectified image. The
axis-aligned bounding boxes are then projected back to the original image.  

Although more complex, the rectification-based approach
allows for the detection of objects with strong perspective distortion in the original
image. However, this comes with the risk of potential spurious detections when
the rectification creates visual artifacts. Experiments show that each
method separately achieves comparable localization performance, and the fusion
of the two reaches the best results.

Both methods rely on vanishing points derived from line segments. %
The points are estimated using sequential
RANSAC~\cite{fischler1981random}: a sample of two line segments defines a vanishing
point. The inliers are segments which angle to the estimated vanishing direction is below a threshold. 

\subsubsection{Vanishing Point Aligned Boxes}
\label{method:vp_alignment}

When the object detection is performed on the original image, the resulting boundary boxes are axis-aligned and they need to be adjusted to obtain perspective boxes.
Since the detection is limited to planar
objects, there is always a plane on which the object lies. The orthogonal box
is updated so that its sides lay on a line incident to the vanishing point of that plane.
To associate the box with these vanishing directions, the following voting
scheme is used: the line segments around the box each vote for the vanishing
direction they support. The two orthogonal vanishing directions with the maximum votes are
used to adjust the box. The adjustment is illustrated in
\Fig~\ref{fig:method_summary}-right: for each box edge, a line between the
midpoint and the vanishing point with the same orientation is derived. The new
box corners are the four intersections of these lines.

\subsubsection{Image Rectification}
\label{method:image_rectification}

The rectifying homography transformation is a composition of two transformations, projective and affine.
The projective transformation sends vanishing points to infinity in horizontal and vertical directions respectively.
The affine transformation is restricted to anisotropic scaling and translation so that it leaves horizontal and vertical directions unchanged.
The affine transformation minimizes the sum of squared differences between the coordinates of line segment endpoints
between the rectified and original images. This choice reduces the amount of re-sampling when the rectified image is rendered.

In the presence of multiple facades, which is when multiple horizontal vanishing points are detected, the image is segmented into planes corresponding to those horizontal vanishing points.
We treat the problem as a 1D classification over the image columns %
where each column is assigned to one of the horizontal vanishing points.
Each horizontal line segment votes for the horizontal vanishing point it supports. The vote is counted over all the columns the line segment intersects. 
The value of the vote is the confidence of the line segment to be associated with the supported vanishing point. This confidence is derived from the softmax over the geometric residuals of this line segment with all the vanishing points. 
Once the histogram of votes is produced for all horizontal vanishing points, the columns are classified using confident majority votes.
Given a column, the majority vote of it is confident when it exceeds a certain threshold and, at the same time, the second best vote is far behind (first to second best ratio). This produces intervals associated with a vanishing point.
Unlabelled column intervals are split by a single threshold between the neighboring labels. The threshold is selected by a maximum likelihood principle.

Each plane is rectified separately before it is processed by the object
detector. The outputted boxes are back-projected to the original image using the inverse
rectification transform.

\subsection{Object Matching}
\label{method:object_matching}

In the object matching step, a set of homographies maximizing the number of corresponding box detections between the two images is found. The following greedy approach is adopted.
For each object in the first image, the $K$ most similar object boxes in the second image are selected. The similarity is measured by the similarity of deep features computed over the boxes. 
A homography hypothesis that maps the object in the first image to the box in the second image is constructed for all $K$ boxes in the second image.
For each hypothesis, all boxes from one image are projected to the second image. If the area \gls{iou} of a projected box and an object box detected in a second image
is greater than a threshold $\varepsilon_{\gls{iou}}=0.5$, the pair is considered corresponding. No visual similarity is enforced on corresponding boxes.
A homography with the highest number of supporting boxes is stored and the supporting boxes are excluded. The procedure is repeated until there is no homography supported by at least two pairs of corresponding boxes.

\subsection{Geometry-Guided Feature Matching}
\label{method:guided_feature_matching}

The object matching step outputs a set of homographies and pairs of object boxes consistent with each homography. 
These boxes provide information about the spatial support of the homography: a
feature in this support complies with this transformation and is guided with
it. A feature in the first image is projected to the second one and matched to
the keypoint with the most similar visual descriptor that falls within a radius
$r_{search}$ from the projection.

\subsection{Additional Feature Matches}
\label{method:match_fusion}

To cover areas with no objects detected or areas off the planes with objects, additional feature matches are added.
This step performs standard feature matching based on the similarity of the feature descriptors.
Local features already matched in the previous step are not considered in this step.

\begin{table*}[tbh]
\centering
\setlength{\tabcolsep}{4pt}
\resizebox{18cm}{!} {
    \begin{tabular}{l*{10}{c}}
\toprule
      \multirow{3}{*}{Method}     & Aachen v1.0                         & Aachen v1.1                         & \multicolumn{6}{c}{RobotCar Seasons} \\
      \cmidrule(r){2-2}
      \cmidrule(r){3-3}
      \cmidrule(r){4-9}
                                  &                                     &                               & \multicolumn{2}{c}{Left}                          & \multicolumn{2}{c}{Rear}                          & \multicolumn{2}{c}{Right} \\
      \cmidrule(r){4-5}
      \cmidrule(r){6-7}
      \cmidrule(r){8-9}
                                  &   Night                             & Night                         & Day                 & Night                       & Day               & Night                         & Day                 & Night \\
      \hline                      
      UprightRootSIFT + NN        & 57.1 / 69.4 / 77.6            & 52.9 / 65.4 / 74.9                  & 51.3 / 76.8 / 90.7  & 4.7 / 9.6 / 16.3            & 57.4 / 81.8 / 96.2& 14.1 / 22.8 / 31.3            & 51.2 / 76.5 / 90.0  & 9.5 / 15.4 / 20.2  \\
      UprightRootSIFT+\horus      & \textbf{74.5 / 84.7 / 98.0}   & \textbf{63.4 / 80.6 / 94.2}         & 51.5 / 76.6 / 90.5  & \textbf{5.0 / 10.8 / 17.8}  & 57.4 / 81.9 / 96.3& \textbf{16.4 / 27.6 / 36.7}   & 51.3 / 76.6 / 90.0  & \textbf{10.3 / 19.0 / 24.7} \\
      \hline                  
      SuperPoint + NN$^{\dagger}$ & 73.5 / 79.6 / 88.8            & $-$ & $-$ & $-$ & $-$ & $-$  & $-$ & $-$\\
      SuperPoint + NN             & 75.5 / 82.7 / 91.8            & 69.1 / 84.8 / 94.8                  & 51.5 / 78.1 / 92.7  &5.9 / 12.6 / 21.1            & 56.7 / 81.8 / 96.2& 20.4 / 42.1 / 63.2            & 52.4 / 79.1 / 93.3  & 12.0 / 24.5 / 37.7   \\
      SuperPoint + \horus         & \textbf{77.6 / 84.7 / 95.9}   & \textbf{69.1 / 85.3 / 96.9}         & 51.3 / 78.0 / 92.6  &5.6 / 13.3 / 23.2            & 56.6 / 81.8 / 96.6& \textbf{21.2 / 42.9 / 64.4}   & 52.6 / 79.0 / 93.2  & \textbf{13.0 / 24.6 / 39.2} \\
      \hline                  
      D2-Net + NN$^{\dagger}$     & 74.5 / 86.7 / 100.0           & $-$ & $-$ & $-$                                                                         & 54.5 / 80.0 / 95.3& 20.4 / 40.1 / 55.0& $-$ & $-$\\
      D2-Net + NN                 & \textbf{79.6 / 89.8 / 100.0}  & \textbf{68.1 / 84.8 / 96.9}         & 52.9 / 80.5 / 95.2  & 15.7 / 37.4 / 56.0          & 54.8 / 81.1 / 96.3& 32.7 / 66.3 / 89.9            & 53.5 / 80.2 / 95.3  & 19.9 / 42.7 / 64.5   \\
      D2-Net + \horus             & \textbf{79.6 / 89.8 / 100.0}  & 67.5 / 84.8 / 97.9                  & 53.3 / 80.2 / 95.0  & \textbf{16.5 / 36.2 / 56.0} & 54.8 / 81.1 / 96.3& \textbf{33.9 / 65.0 / 88.8}   & 53.3 / 80.1 / 95.1  & \textbf{21.0 / 43.2 / 63.4} \\
      \bottomrule	
    \end{tabular}
  }
  \caption{Comparison with NN matching on \gls{sota} local features. Evaluation metric is the percentage of images whose registration error in translation and rotation falls below  (0.5m, 2$^{\circ}$) / (1m, 5$^{\circ}$) / (5m,  10$^{\circ}$) respectively. 
  $^{\dagger}$ denotes results published in the original papers, other results are obtained by our execution.}
  \label{tab:horus_boost}
\end{table*}

\section{Implementation Details}

\PAR{Object Detection Training.}
Objects are detected using the Faster R-CNN network~\cite{ren2015faster} with a
Resnet-50~\cite{he2016deep} backbone with Tensorflow~\cite{abadi2016tensorflow}.
The network is trained in three steps adopted from the authors'
guidelines~\cite{ren2015faster}. At test time, objects are
detected at multiple scales of the image ($\times$1 and $\times$2).
For training, we use the box annotations of the window instances in the
\gls{oid}~\cite{kuznetsova2020open}.
This dataset holds a significant amount of noisy labels that hinder the
precision and the generalization of the network. So images with such labels are
discarded (\example incomplete labelling, occluded windows).  This process keeps
only 10\% of the images to which are added images from the CMP facades
dataset~\cite{Tylecek13}. The final dataset holds 9278 images and 117632 boxes.

\PAR{Object Pair Homography.} A pair of corresponding objects
provides four point-to-point correspondences, which is enough to instantiate a homography.
The homography used for guided matching of the local features is estimated from all
box correspondences that support the initial homography. A vanishing point consistency is
enforced in this step.

\PAR{Line Segment Detection.}
Line segments are detected with the augmented HT-LCNN~\cite{lin2020deep} and
segments smaller than 20 pixels are discarded. 

\section{Evaluation}

This section provides an evaluation of \horus by comparison to baseline methods, an ablation study, and a comparison to different \gls{sota} approaches. 

\subsection{Experimental Setup}
\horus is evaluated against other feature matching methods: the default
\gls{nn} approach, the coarse-to-fine Patch2Pix~\cite{zhou2021patch2pix} and
LOFTR~\cite{sun2021loftr}, the graph-based approach
SuperGlue~\cite{sarlin2020superglue}, and the filtering method
AdaLAM~\cite{cavalli2020handcrafted}. The evaluation measures the pose
estimation performance for outdoor localization on three benchmark datasets:
Aachen v1.0, Aachen V1.1, RobotCar Seasons.

\PAR{Visual Localization.} We follow the standard evaluation setup from the
localization
benchmark~\footnote{https://www.visuallocalization.net/benchmark/}. The
released code takes feature matches as input and runs 3D structure-based
localization relying on the COLMAP
library~\cite{schonberger2016structure,schoenberger2016mvs}. The localization
is evaluated with the percentage of estimated poses within an error threshold
with respect to the groundtruth. The same rotation and translation thresholds
as in the benchmark are used here: (0.25m, 2$^{\circ}$) / (0.5m, 5$^{\circ}$) /
(5m, 10$^{\circ}$).

\PAR{Datasets.} Experiments are run on three standard urban localization
datasets: the Aachen
datasets (v1.0 and v1.1)~\cite{sattler2018benchmarking,sattler2012image,zhang2021reference}
and the RobotCarSeasons one~\cite{maddern20171,sattler2018benchmarking}.  The
last one presents additional challenges because of the lower image quality and
the presence of artefacts such as motion blur, raindrops, and overexposure.
For Aachen, the list of image pairs to match is provided by
the benchmark. For the other dataset, no list is provided so we use
the ground-truth positions to derive reference image pairs and use the global
image descriptor denseVLAD~\cite{torii201524} to match query images to the 20 most
similar reference ones, as in~\cite{dusmanu2019d2}.

\begin{figure*}[thb]
 \centering
 \includegraphics[width=0.66\linewidth]{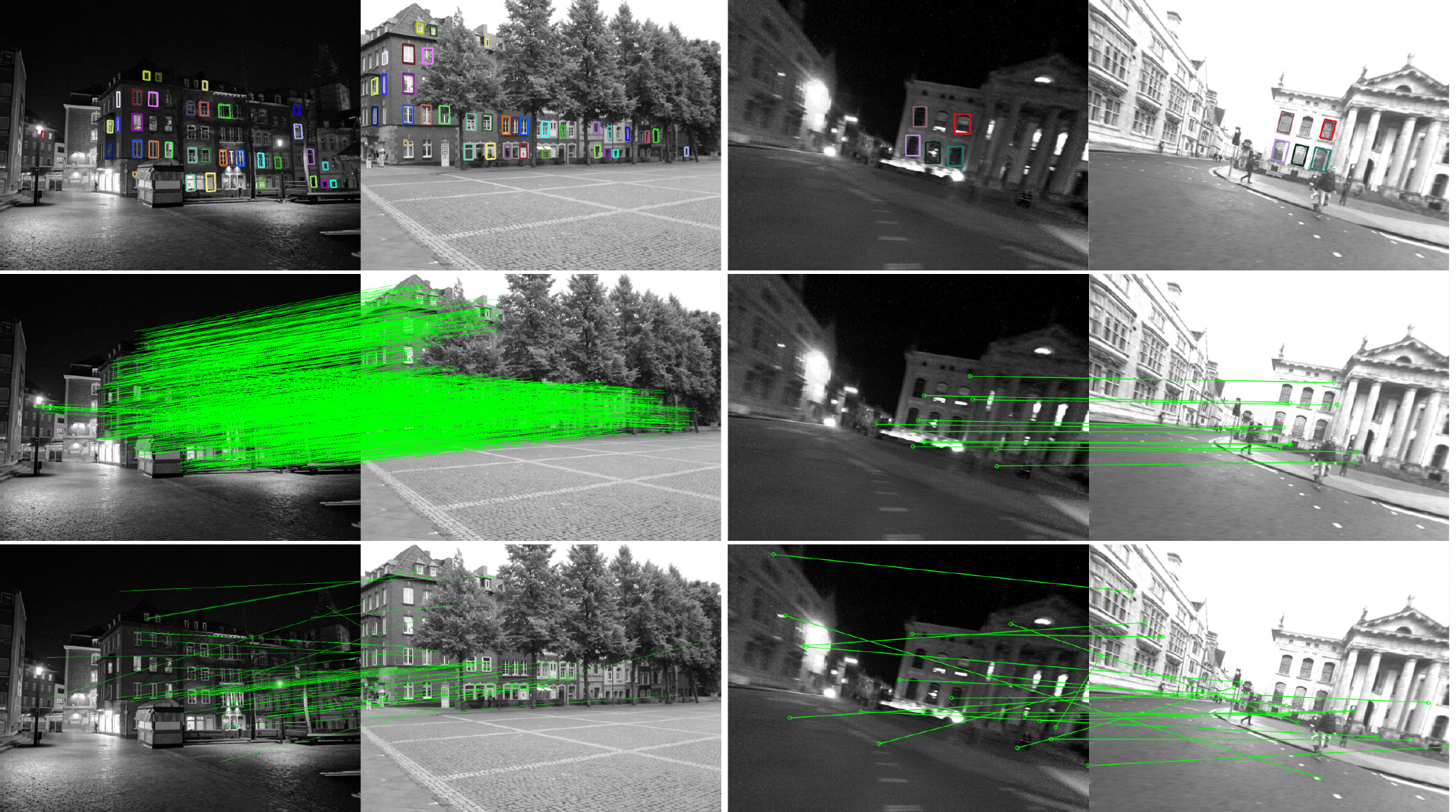}
 \includegraphics[width=0.33\linewidth]{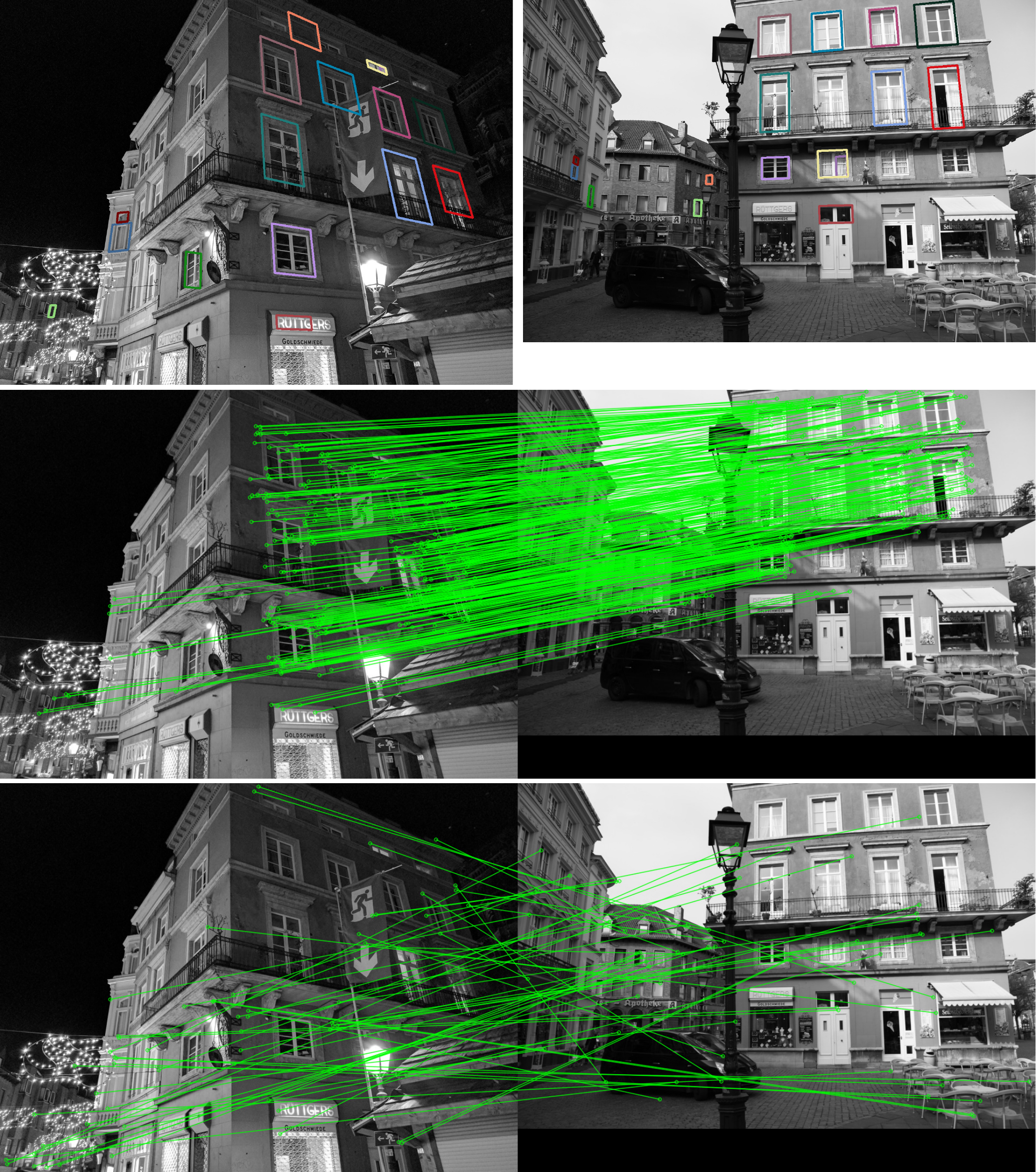}
  \caption{Qualitative results. 
  Top-Down: Box matches, \horus guided matches, \gls{nn} matches. 
  Left-Right: SIFT on Aachen v1.0, SIFT on RobotCarSeasons, D2-Net on Aachen
  v1.0.
  The examples show that the box detection and matching are robust to strong
  illumination variations and occlusions. This allows the guided matching of
  features in situations where \gls{nn} fails.}
 \label{fig:success_fail_cases}
\end{figure*}

\subsection{Comparison to baseline methods}

In \Tab~\ref{tab:horus_boost}, we compare the method against \gls{nn} matching for three local features: the
hand-crafted upright-root-SIFT~\cite{arandjelovic2012three,lowe2004distinctive}
and two \gls{sota} deep features SuperPoint~\cite{detone18superpoint} and
D2-Net~\cite{dusmanu2019d2}. The SIFT features are provided with the
datasets and the deep features are extracted with the author's code with the
parameters reported in the benchmark. Except for the input matches, the
localization pipeline stays fixed.

On the Aachen datasets, the \horus matches consistently improve the performance
over the baseline matches for SIFT by up to 17\%. Qualitative results using SIFT features are shown in \Fig~\ref{fig:success_fail_cases}.
The improvement deteriorates with more advanced features such as
with D2-Net and SuperPoint. Even though the quantitative results are on par, the set of resolved images is different, see~\Fig\ref{fig:success_fail_cases}-right, which shows the potential of the proposed method.

On the RobotCar Seasons, \horus increases the recall for night queries while
preserving the scores on the day queries. It also allows registering more night
queries, between 50 and 100.
This shows that \horus is a relevant method for feature
matching across day-night images.
The bottleneck for the RobotCar improvement is unreliable box detection.
This happens when the image does not exhibit windows, when the car goes along a
road, or when the image is too deteriorated (blur, overexposure).

\subsection{Ablation study}

We assess the advantage of object-guided feature
matching and how each element of the method contributes to the performances.

\begin{table}[tbh]
  \centering
    \begin{tabular}{lcc}
      \toprule
                        & without AF                          &  with AF \\
	    \midrule
      O                 & 44.9 / 60.2 / 76.5          & 71.4 / 82.7 / 94.9 \\
      OA                & 42.9 / 60.2 / 78.6          & 70.4 / 84.7 / 95.9 \\
	    \midrule
      R                 & 41.8 / 59.2 / 76.5          & 72.4 / 82.7 / 93.9 \\ %
      RA                & 45.9 / 63.3 / 77.6          & 70.4 / 82.7 / 93.9 \\ %
	    \midrule
      O + R             & \hblue{\textbf{60.2 / 73.5 / 88.8}} & 71.4 / 86.7 / 98.0 \\ 
      (O + R)$\cdot$A   & 59.2 / 76.5 / 88.8          & \hgreen{\textbf{73.5 / 85.7 / 96.9}} \\ %
      \bottomrule	
    \end{tabular}
  \caption{Comparison of the perspective box detection on Aachen v1.0.  O:
  Orthogonal boxes. OA: Orthogonal boxes Adjusted. R: orthogonal boxes on
  Rectified images. RA: boxes on Rectified images Adjusted. With / without additional features (AF).
  \hblue{\textbf{Blue}}: Best without AF.
  \hgreen{\textbf{Green}}: Best with AF.
  }
\label{tab:results_box_ablation}
\end{table}

\paragraph{Perspective Box Detection (\Tab~\ref{tab:results_box_ablation})}
We evaluate the influence of the box derivation on the localization performance
using the SIFT features on Aachen v1.0. Results show that both derivations
achieve comparable results on their own. When no additional features (Sec. \ref{method:match_fusion}) are used,
merging and adjusting the boxes on the original and rectified images
significantly improves the scores. One explanation is that these two
derivations detect more boxes together.

\paragraph{Plane segmentation.}
Perspective box detection on the rectified images uses plane segmentation to rectify only the part of the image relevant to the vanishing point.
This task has been previously addressed in~\cite{wan2011automatic}, introducing a cost associated with plane transitions (smoothness cost). In our experiments,
segmentation of~\cite{wan2011automatic} produces, for the best value of the cost mixing parameter, 7\% worse results than our method on Aachen v1.0. We conclude,
that for challenging images, it is difficult to set the relative weight of the smoothness cost.

\paragraph{Constrained Box Matching (\Tab~\ref{tab:ablation_vp_constraint})} The
coarse geometry derived from matching objects leads to better results
when constrained by the vanishing directions (+15-20\%). This is emphasised
when no additional features are used.
Using the vanishing directions to adjust the boxes also improves the results.
This suggests that integrating the vanishing points in the method is relevant
to compensate for the inaccurate box corners. 

\begin{table}[tbh]
  \centering
  \setlength{\tabcolsep}{4pt}
    \begin{tabular}{lcc}
      \toprule
                                                & without AF                                 & with AF \\
	    \midrule
      C + A                                   & 59.2 / 76.5 / 88.8                    & \hgreen{\textbf{73.5 / 85.7 / 96.9}} \\
      C + $\bar{\rm A}$                       & \hblue{\textbf{60.2 / 73.5 / 88.8}}  & 71.4 / 86.7 / 98.0 \\ 
	    \midrule
      $\overline{\rm C}$ + A                  & 45.9 / 62.2 / 77.6                    & 69.4 / 84.7 / 95.9 \\
      $\overline{\rm C}$ + $\overline{\rm A}$ & 40.8 / 55.1 / 72.4                    & 68.4 / 84.7 / 94.9 \\
      \bottomrule
    \end{tabular}
  \caption{Influence of the
  \textbf{C}onstrained box matching with box \textbf{A}djustement (A) and
  without. \hblue{\textbf{Blue}}: Best without AF.  \hgreen{\textbf{Green}}: Best
  with AF.}
  \label{tab:ablation_vp_constraint}
\end{table}

\paragraph{Box Description (\Tab~\ref{tab:ablation_box_description})}
The aggregated D2-Net descriptors achieve better results than the detector's
description. One explanation may be that the detector's features are optimized
for object classification which may not be suited for matching. Note that when
additional features are used, the method is not as sensitive to the box description
as without.

\begin{table}[tbh]
  \centering
    \begin{tabular}{lcc}
      \toprule
                                                        & without AF                                 & with AF \\
	    \midrule
      D2-Net + GeM.                                     & \hblue{\textbf{55.1 / 72.4 / 85.7}} & \hgreen{\textbf{73.5 / 83.7 / 95.9}} \\
      GeM.                                              & 42.9 / 57.1 / 68.4                  & 70.4 / 84.7 / 94.9 \\
      Avg\_pool                                         & 38.8 / 53.1 / 69.4                  & 67.3 / 79.6 / 93.9 \\ 
      \bottomrule	
    \end{tabular}
  \caption{Comparison of the box description used to prune candidate box
  matches on Aachen v1.0. \hblue{\textbf{Blue}}: Best without AF.
  \hgreen{\textbf{Green}}: Best with AF.  }
\label{tab:ablation_box_description}
\end{table}

\paragraph{Parameter Sensitivity.}
Additional experiments show that \horus is not very sensitive to the
feature search radius, the number of candidate match for
one box. The method seems to be robust to noisy box detection as it performs
similarly even with low confidence detections. One reason for this may be the
robust estimation of the box matching.

\subsection{Comparison with the \gls{sota}}

\begin{figure}[thb]
 \centering
 \includegraphics[width=\linewidth]{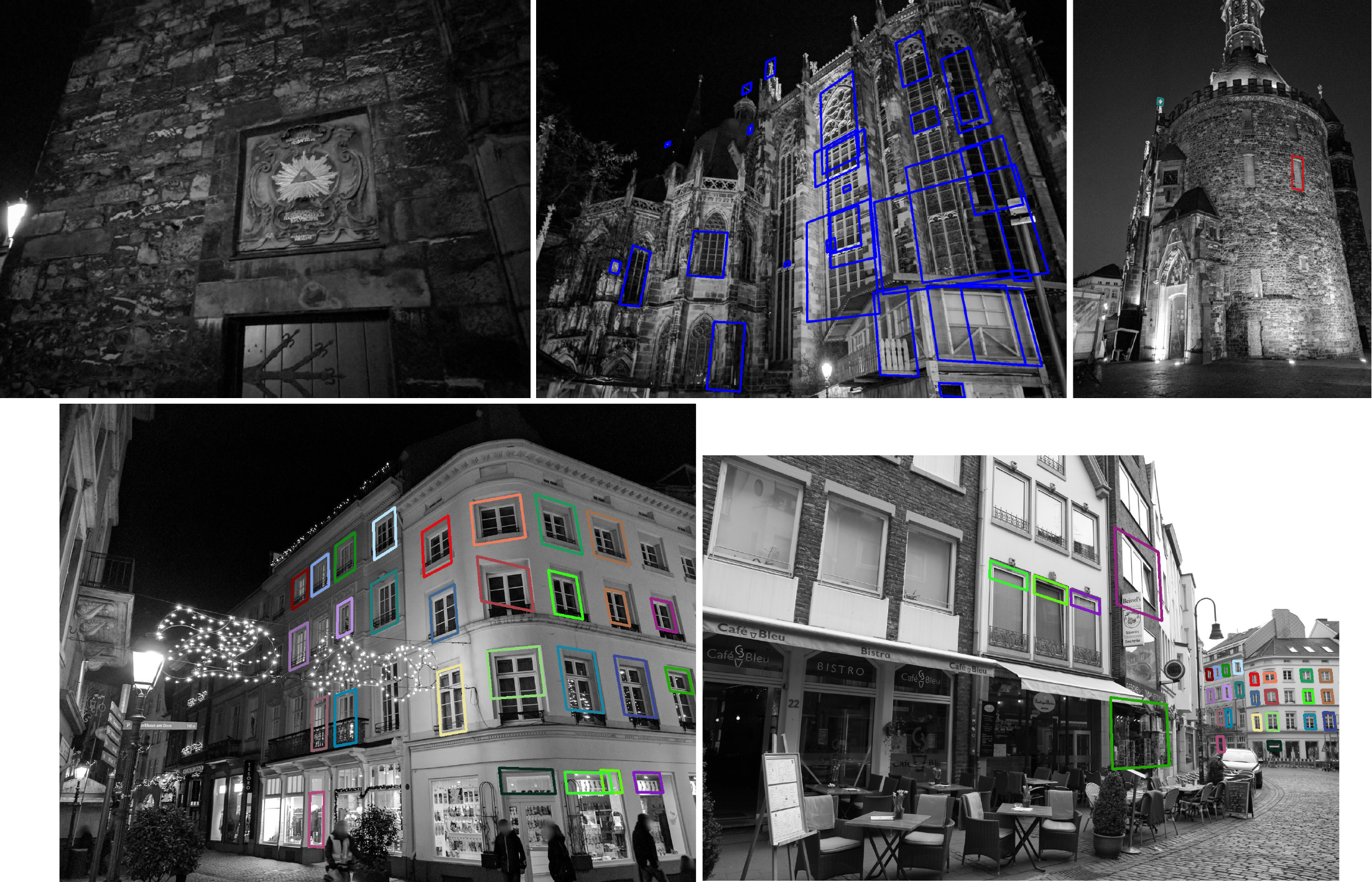}
  \caption{Failure cases for object detection (top) and box matching (bottom).}
 \label{fig:failure-cases}
\end{figure}

Results in \Tab~\ref{tab:soa} show that our method 
is competitive with \gls{sota} matching approaches. One
advantage of our method (and Patch2Pix) is that it is agnostic to the local
features whereas other existing approaches, such as LOFTR or SuperGlue, train
the guided matching for a specific feature. Although the numerical
performances are comparable, the successful matching pairs of \horus and other
methods are complementary (\Fig~\ref{fig:success_fail_cases}-right). This shows that
object-guided matching is a relevant method to push the current limits of feature
matching.
Failure cases of the proposed method are shown in \Fig\ref{fig:failure-cases}:
obvious limitations of the method include images with no objects to detect,
incorrect box detection, and noise in the object matching.

\begin{table}[tbh]
\centering
\setlength{\tabcolsep}{4pt}
\resizebox{8.5cm}{!} {
    \begin{tabular}{lcc}
      \toprule
      & Aachen v1.0 & Aachen v1.1 \\
      \midrule
      SIFT + \horus                                                  & 74.5 / 84.7 / 98.0           & 63.4 / 80.6 / 94.2 \\
      SuperPoint + \horus                                            & 77.6 / 84.7 / 95.9           & \textbf{69.1 / 85.3 / 96.9}  \\
      D2Net + \horus                                                 & \textbf{79.6 / 89.8 / 100.0} & 67.5 / 84.8 / 97.9\\
      \midrule
      SIFT + AdaLAM~\cite{cavalli2020handcrafted}                   & 69.4 / 81.6 / 92.9            & $-$\\ %
      Patch2Pix$^{\dagger}$~\cite{zhou2021patch2pix}                & 78.6 / 88.8 / 99.0            & $-$\\
      LOFTR$^{\dagger}$~\cite{sun2021loftr}                         & $-$                           & 72.8 / 88.5 / 99.0 \\
      SuperPoint + SuperGlue$^{\dagger}$~\cite{sarlin2020superglue} & \textbf{79.6 / 90.8 / 100.0}  & \textbf{73.3 / 88.0 / 98.4} \\ %
      \bottomrule
    \end{tabular}
  }
  \caption{Localization Performance on Aachen. $^{\dagger}$ means we report the results
  from the paper.}
  \label{tab:soa}
\end{table}

\section{Conclusion}

 Object-Guided Localization (\horus) for urban scenes was introduced. The approach is based on a novel method of local-feature matching.
 The proposed method overcomes significant changes in illumination by first detecting semantic objects and establishing their correspondence between images.
 Object correspondences provide hypotheses of planar homography that are used to guide the matching of local features.
 
 We have experimentally shown the potential of the method. Significant improvements were achieved with basic SIFT features. With the \gls{sota} D2Net features, the method is on par or slightly better than the standard approach, resolving and failing in different cases, demonstrating the complementarity of the two approaches.

\section{Acknowledgements}
The authors thank Tom{\'a}\v{s} Jen{\'i}\v{c}ek and Alan Luke\v{z}i\v{c} for
their thoughtful comments.

\bibliographystyle{ieee_fullname}  %
\bibliography{main}

\end{document}